\documentclass[11pt,a4paper]{article}
\usepackage[hyperref]{acl2021}
\usepackage{times}
\usepackage{latexsym}

\usepackage{microtype}

\usepackage{amsfonts}
\usepackage{amsmath,amssymb,bm}
\usepackage{graphicx}
\usepackage{multirow}

\usepackage{tabularx}
\newcolumntype{C}[1]{>{\centering\arraybackslash}p{#1}}

\usepackage{algorithm}
\usepackage[noend]{algorithmic}

\aclfinalcopy 

\title{A Span-Based Model for Joint Overlapped and Discontinuous\\ Named Entity Recognition}

\author{
Fei Li$^1$ \and Zhichao Lin$^{2}$ \and Meishan Zhang$^{2}$ \and Donghong Ji$^1$\thanks{~~Corresponding author.} \\
1. Department of Key Laboratory of Aerospace Information Security and Trusted Computing,\\
Ministry of Education, School of Cyber Science and Engineering, Wuhan University, China \\ 
2. School of New Media and Communication, Tianjin University, China \\
\texttt{\{lifei\_csnlp,dhji\}@whu.edu.cn} \and \texttt{mason.zms@gmail.com} \\
}

\date{}

\begin{document}
\maketitle
\begin{abstract}
Research on overlapped and discontinuous named entity recognition (NER) has received increasing attention.
The majority of previous work focuses on either overlapped or discontinuous entities.
In this paper, we propose a novel span-based model that can recognize both overlapped and discontinuous entities jointly.
The model includes two major steps.
First, entity fragments are recognized by traversing over all possible text spans,
thus, overlapped entities can be recognized.
Second, we perform relation classification to judge whether a given pair of entity fragments to be overlapping or succession.
In this way, we can recognize not only discontinuous entities, and meanwhile doubly check the overlapped entities.
As a whole, our model can be regarded as a relation extraction paradigm essentially.
Experimental results on multiple benchmark datasets (i.e., CLEF, GENIA and ACE05) show that our model is highly competitive for overlapped and discontinuous NER. 
\end{abstract}

\section{Introduction}

\begin{figure*}[t]
\centering
\includegraphics[width=0.95\textwidth]{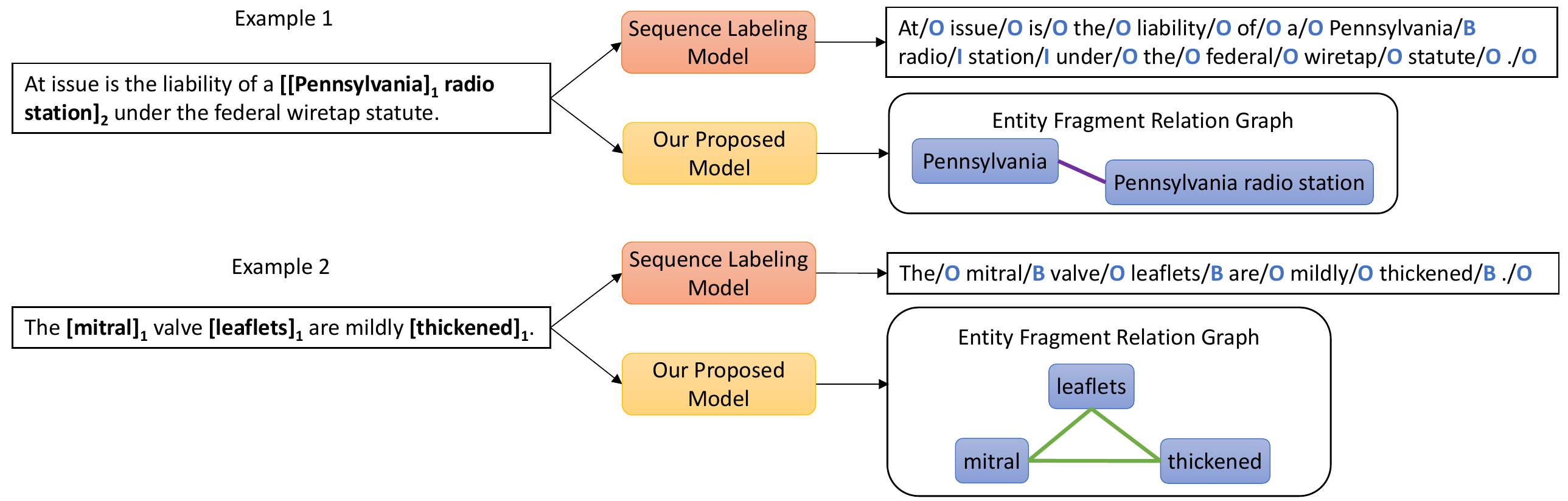}
\caption{Examples to illustrate the differences between the sequence labeling model and our span-based model.
On the left, word fragments marked with the same number belong the same entity. On the right, blue rectangles denote the recognized entity fragments, and solid lines indicate the \texttt{Succession} or \texttt{Overlapping} relations between them (the two relations are mutually exclusive).
}
\label{fig:example}
\end{figure*}

Named entity recognition (NER)~\cite{sang2003introduction} is one fundamental task for natural language processing (NLP), due to its wide application in information extraction and data mining~\cite{lin2019reliability,cao2019low}. Traditionally, NER is presented as a sequence labeling problem and widely solved by conditional random field (CRF) based models~\cite{lafferty2001conditional}.
However, this framework is difficult to handle overlapped and discontinuous entities~\cite{lu2015joint,muis2016learning}, which we illustrate using two examples as shown in Figure~\ref{fig:example}.
The two entities ``Pennsylvania'' and ``Pennsylvania radio station'' are nested with each other,\footnote{
We consider ``nested'' as a special case of ``overlapped''.}
and the second example shows a discontinuous entity ``mitral leaflets thickened'' involving three fragments.

There have been several studies to investigate overlapped or discontinuous entities~\cite{finkel2009nested,lu2015joint,muis2017labeling,katiyar2018nested,wang2018neural,ju2018neural,wang2018transition,fisher2019merge,luan2019general,wang2019combining}. The majority of them focus on overlapped NER, with only several exceptions to the best of our knowledge.
\citet{muis2016learning} present a hypergraph model that is capable of handling both overlapped and discontinuous entities.
\citet{wang2019combining} extend the hypergraph model with long short-term memories (LSTMs)~\cite{hochreiter1997long}. \citet{dai-etal-2020-effective} proposed a transition-based neural model for discontinuous NER.
By using these models, NER could be conducted universally without any assumption to
exclude overlapped or discontinuous entities, which could be more practical in real applications.

The hypergraph \cite{muis2016learning,wang2019combining} and transition-based models \cite{dai-etal-2020-effective} are flexible to be adapted for different tasks, achieving great successes for overlapped or discontinuous NER.
However, these models need to manually define graph nodes, edges and transition actions. Moreover, these models build graphs or generate transitions along the words in the sentences gradually, which may lead to error propagation \cite{zhang2016transition}.
In contrast, the span-based scheme might be a good alternative,
which is much simpler including only span-level classification.
Thus, it needs less manual intervention and meanwhile span-level classification can be fully parallelized without error propagation.
Recently, \citet{luan2019general} utilized the span-based model
for information extraction effectively.

In this work, we propose a novel span-based joint model to recognize overlapped and discontinuous entities simultaneously in an end-to-end way.
The model utilizes BERT~\cite{devlin-etal-2019-bert} to produce deep contextualized word representations,
and then enumerates all candidate text spans ~\cite{luan2019general},
classifying whether they are entity fragments.
Following, fragment relations are predicted by another classifier to determine whether two specific fragments involve a certain relation.
We define two relations for our goal: \texttt{Overlapping} or \texttt{Succession},
which are used for overlapped and discontinuous entities, respectively.
In essence, the joint model can be regarded as one kind of relation extraction models,
which is adapted for our goal.
To enhance our model, we utilize the syntax information as well by using a dependency-guided graph convolutional network~\cite{kipf2017semi,zhang2018graph,jie2019dependency,guo2019attention}.

We evaluate our proposed model on several benchmark datasets which includes both overlapped and discontinuous entities (e.g., CLEF~\cite{suominen2013overview}).
The results show that our model outperforms the hypergraph \cite{muis2016learning,wang2019combining} and transition-based models \cite{dai-etal-2020-effective}.
Besides, we conduct experiments on two benchmark datasets including only overlapped entities (i.e., GENIA~\cite{kim2003genia} and ACE05).
Experimental results show that our model can also obtain comparable performances with the state-of-the-art models~\cite{luan2019general,wadden2019entity,strakova2019neural}.
In addition, we observe that our approaches for model enhancement are effective in the benchmark datasets.
Our code is available at \url{https://github.com/foxlf823/sodner}.

\section{Related Work}
In the NLP domain, NER is usually considered as a sequence labeling problem~\cite{liu2018empower,lin2019reliability,cao2019low}.
With well-designed features, CRF-based models have achieved the leading performance~\cite{lafferty2001conditional,finkel2005incorporating,liu2011recognizing}.
Recently, neural network models have been exploited for feature representations~\cite{chen2014fast,zhou2015neural}.
Moreover, contextualized word representations such as ELMo~\cite{peters2018deep}, Flair~\cite{akbik2018contextual} and BERT~\cite{devlin-etal-2019-bert} have also achieved great success.
As for NER, the end-to-end bi-directional LSTM CRF models~\cite{lample-EtAl:2016:N16-1,ma-hovy:2016:P16-1,yang2018design} is one representative architecture.
These models are only capable of recognizing regular named entities.

\begin{figure*}[t]
\centering
\includegraphics[width=0.90\textwidth]{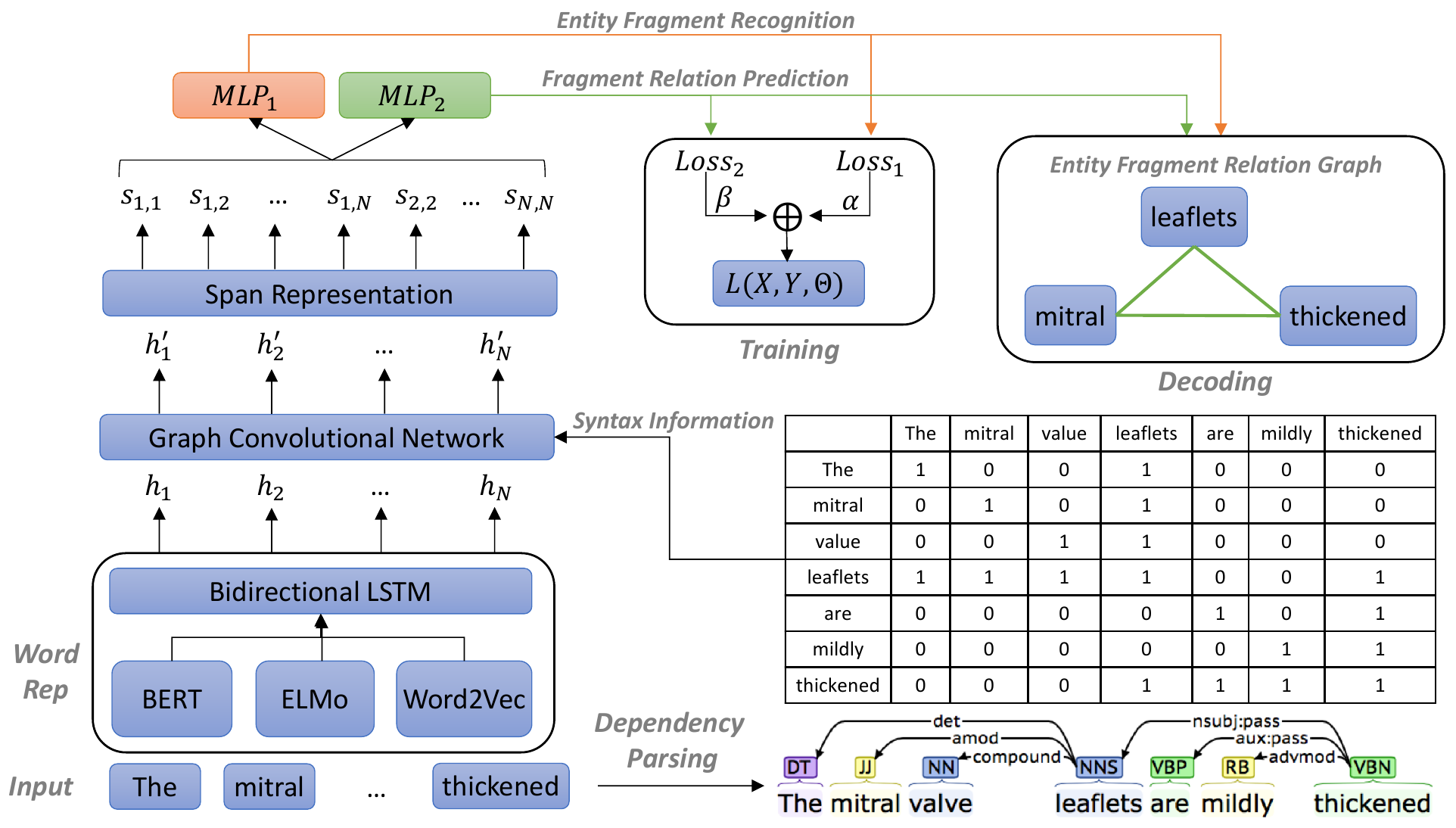}
\caption{The architecture of our model. The input is ``The $[\textbf{mitral}]_1$ valve $[\textbf{leaflets}]_1$ are mildly $[\textbf{thickened}]_1$''. $\bm{h}_1$ denotes the original word representation and $\bm{h}'_1$ denotes the syntax-enhanced word representation. $\bm{s}_{1,2}$ denotes the span representation. $\alpha$ and $\beta$ control the loss weights of two tasks, namely recognizing entity fragments from text spans and predicting the relation between each pair of fragments.
}
\label{fig:model}
\end{figure*}

For overlapped NER, the earliest model to our knowledge is proposed by \citet{finkel2009nested}, where they convert overlapped NER as a parsing task.
\citet{lu2015joint} propose a hypergraph model to recognize overlapped entities and lead to a number of extensions~\cite{muis2017labeling,katiyar2018nested,wang2018neural}.
Moreover, recurrent neural networks (RNNs) are also used for overlapped NER~\cite{ ju2018neural,wang2018transition}.
Other approaches include multi-grained detection~\cite{xia2019multi}, boundary detection~\cite{zheng2019boundary}, anchor-region network~\cite{lin2019sequence} and machine reading comprehension~\cite{li-etal-2020-unified}.
The state-of-the-art models for overlapped NER include the sequence-to-sequence (seq2seq) model~\cite{strakova2019neural}, where the decoder predicts multiple labels for a word and move to next word until it outputs the ``end of word'' label, and the span-based model~\cite{luan2019general,wadden2019entity}, where overlapped entities are recognized by classification for enumerated spans.

Compared with the number of related work for overlapped NER,
there are no related studies for only discontinuous NER, but several related studies for both overlapped and discontinuous NER.
Early studies addressed such problem by extending the BIO label scheme \cite{tang2013recognizing,metke2016concept}.
\citet{muis2016learning} first proposed a hypergraph-based model for recognizing overlapped and discontinuous entities, and
then \citet{wang2019combining} utilized deep neural networks to enhance the model. 
Very recently, \citet{dai-etal-2020-effective} proposed a transition-based neural model with manually-designed actions for both overlapped and discontinuous NER.
In this work, we also aim to design a competitive model for both overlapped and discontinuous NER. Our differences are that our model is span-based~\cite{luan2019general} and it is also enhanced by dependency-guided graph convolutional network (GCN)~\cite{zhang2018graph,guo2019attention}.

To our knowledge, syntax information is commonly neglected in most previous work for overlapped or discontinuous NER, except \citet{finkel2009nested}.
The work employs a constituency parser to transform a sentence into a nested entity tree, and syntax information is used naturally to facilitate NER.
By contrast, syntax information has been utilized in some studies for traditional regular NER.
Under the traditional statistical setting, syntax information is used by manually-crafted features~\cite{hacioglu2005detection,ling2012fine} or auxiliary tasks~\cite{florian2006factorizing} for NER.
Recently, \citet{jie2017efficient} build a semi-CRF model based on dependency information to optimize the research space of NER recognition.
\citet{jie2019dependency} stack the dependency-guided graph convolutional network~\cite{zhang2018graph,guo2019attention} on top of the BiLSTM layer.
These studies have demonstrated that syntax information could be an effective feature source for NER.

\section{Method}
\label{sec:method}

The key idea of our model includes two mechanisms.
First, our model enumerates all possible text spans in a sentence and then exploits
a multi-classification strategy to determine whether one span is an entity fragment as well as the entity type.
Based on this mechanism, overlapped entities could be recognized.
Second, our model performs pairwise relation classifications over all entity fragments to recognize their relationships.
We define three kinds of relation types:
\begin{itemize}
\setlength{\itemsep}{0pt}
\setlength{\parsep}{0pt}
\setlength{\parskip}{0pt}
  \item \texttt{Succession}, indicating that the two entity fragments belong to one single named entity.
  \item \texttt{Overlapping}, indicating that the two entity fragments have overlapped parts.
  \item \texttt{Other}, indicating that the two entity fragments have other relations or no relations.
\end{itemize}

With the \texttt{Succession} relation, we can recognize discontinuous entities.
Through the \texttt{Overlapping} relation, we aim to improve the recognition of overlapped entities with double supervision.
The proposed model is essentially a relation extraction model being adapted for our task.
The architecture of our model is illustrated in Figure~\ref{fig:model},
where the main components include the following parts: (1) word representation, (2) graph convolutional network, (3) span representation, and (4) joint decoding,
which are introduced by the following subsections, respectively.

\subsection{Word Representation}
\label{sec:word_rep}

We exploit BERT \cite{devlin-etal-2019-bert} as inputs for our model, which has demonstrated effective for a range of NLP tasks.\footnote{We also investigate the effects of different word encoders in the experiments. Please refer to Appendix \ref{app:cmp_wordrep}.}
Given an input sentence $\bm{x}$ = \{$x_1$, $x_2$, ..., $x_N$\}, we convert each word $x_i$ into word pieces and then feed them into a pretrained BERT module.
After the BERT calculation, each sentential word may involve vectorial representations of several pieces.
Here we employ the representation of the beginning word piece as the final word representation following \cite{wadden2019entity}.
For instance, if ``fevers'' is split into ``fever'' and ``\#\#s'', the representation of ``fever'' is used as the whole word representation.
Therefore, all the words in the sentence $\bm{x}$ correspond to a matrix $\bm{H}$ = \{$\bm{h}_1$, $\bm{h}_2$, ..., $\bm{h}_N$\} $\in\mathbb{R}^{N\times d_{h}}$,
where $d_{h}$ denotes the dimension of $\bm{h}_i$.

\subsection{Graph Convolutional Network}
\label{sec:gcn}
Dependency syntax information has been demonstrated to be useful for NER previously~\cite{jie2019dependency}.
In this work,  we also exploit it to enhance our proposed model.\footnote{Some cases are shown in Appendix \ref{app:case_studies}.}
Graph convolutional network (GCN) \cite{kipf2017semi} is one representative method to encode dependency-based graphs,
which has been shown effective in information extraction~\cite{zhang2018graph}.
Thus, we choose it as one standard strategy to enhance our word representations.
Concretely, we utilize the attention-guided GCN (AGGCN)~\cite{guo2019attention} to reach our goal,
as it can bring better performance compared with the standard GCN.

\begin{figure}[t]
\centering
\includegraphics[scale=0.6]{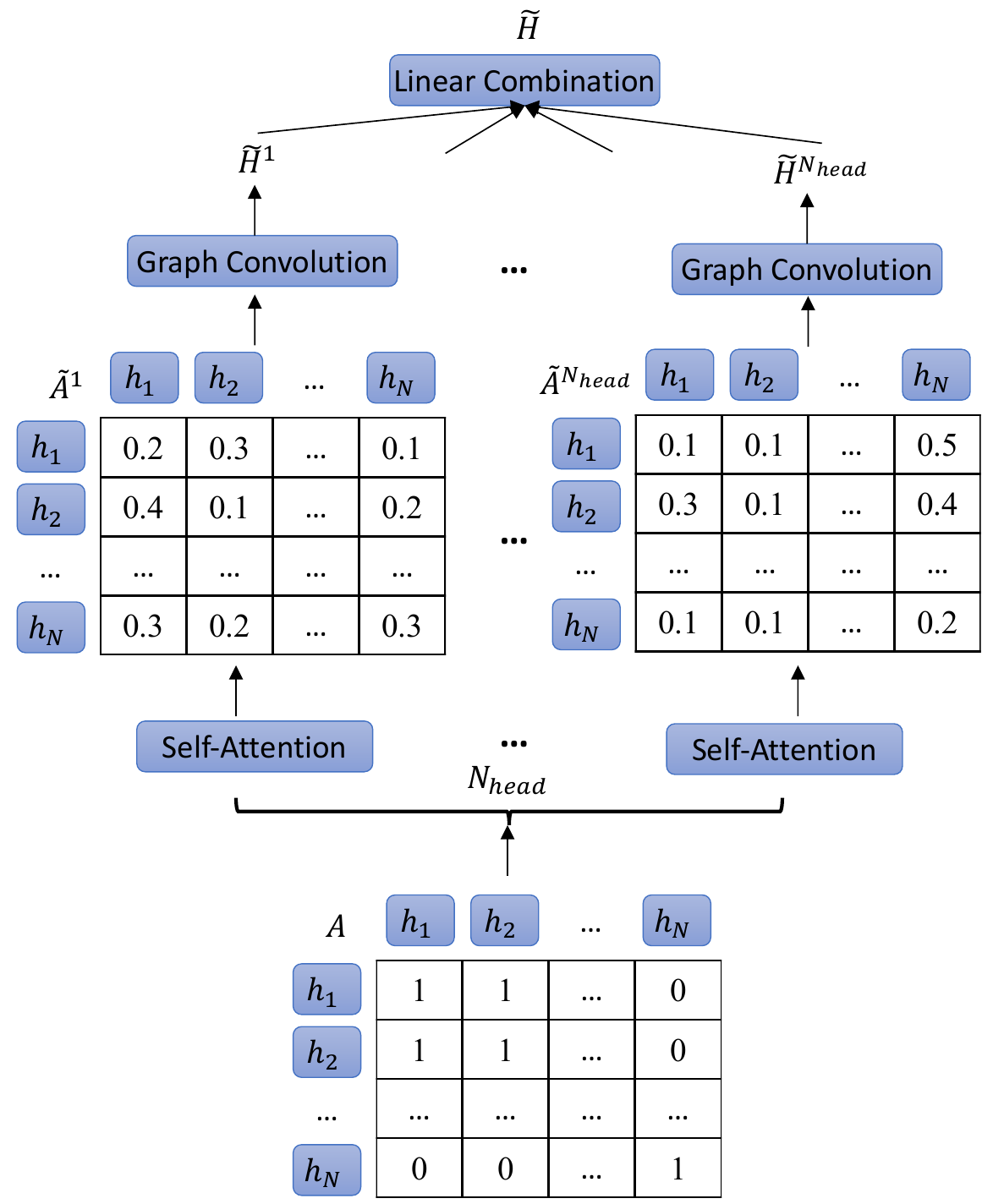}
\caption{\label{fig:aggcn} The architecture of our graph convolutional network. Graph Convolution: Equation \ref{eqn:gcn}. Self-Attention: Equation \ref{eqn:self-att}.
}
\end{figure}

In order to illustrate the network of AGGCN (Figure \ref{fig:aggcn}), we start with the standard GCN module.
Given the word representations $\bm{H}$ = \{$\bm{h}_1$, $\bm{h}_2$, ..., $\bm{h}_N$\},
the standard GCN uses the following equation to update them:
\begin{equation}
\bm{h}_{i}^{(l)} = \sigma (\sum_{j=1}^{N} \bm{A}_{ij} \bm{W}^{(l)} \bm{h}_{j}^{(l-1)} + \bm{b}^{(l)} ),
\label{eqn:gcn}
\end{equation}
where $\bm{W}^{(l)}$ and $\bm{b}^{(l)}$ are the weight and bias of the $l$-th layer.
$\bm{A}$ $\in\mathbb{R}^{N\times N}$ is an adjacency matrix obtained from the dependency graph,
where $\bm{A}_{ij}=1$ indicates there is an edge between the word $i$ and $j$ in the dependency graph.
Figure \ref{fig:model} offers an example of the matrix which is produced by the corresponding dependency syntax tree.

In fact, $\bm{A}$ can be considered as a form of hard attention in GCN, while AGGCN~\cite{guo2019attention} aims to improve the method by using $\bm{A}$ in the lower layers and updating $\bm{A}$ at the higher layers via multi-head self-attention~\cite{vaswani2017attention} as below:
\begin{equation}
\bm{\tilde{A}}^{t} = \mathrm{softmax}(\frac{\bm{H}^{t}\bm{W}^{t}_{Q} \times (\bm{H}^{t}\bm{W}^{t}_{K})^{T}}{\sqrt{d_{head}}}),
\label{eqn:self-att}
\end{equation}
where $\bm{W}^{t}_{Q}$ and $\bm{W}^{t}_{K}$ are used to project the input $\bm{H}^{t}$ $\in\mathbb{R}^{N\times d_{head}}$ ($d_{head}=\frac{d_h}{N_{head}}$) of the $t$-th head into a query and a key. $\bm{\tilde{A}}^{t}$ $\in\mathbb{R}^{N\times N}$ is the updated adjacency matrix for the $t$-th head.

For each head $t$, AGGCN uses $\bm{\tilde{A}}^{t}$ and a densely connected layer to update the word representations, 
which is similar to the standard GCN as shown in Equation~\ref{eqn:gcn}. The output of the densely connected layer is  $\bm{\tilde{H}}^{t}$ $\in\mathbb{R}^{N\times d_{h}}$.
Then a linear combination layer is used to merge the output of each head, namely $\bm{\tilde{H}} = [\bm{\tilde{H}}^{1}, \cdots ,\bm{\tilde{H}}^{N_{head}}] \bm{W}_{1}$, where $\bm{W}_{1}$ $\in\mathbb{R}^{(N_{head}\times d_{h})\times d_{h}}$ is the weight and $\bm{\tilde{H}}$ $\in\mathbb{R}^{N\times d_{h}}$ is the final output of AGGCN.

After that, $\bm{\tilde{H}}$ is concatenated with the original word representations $\bm{H}$ to form final word representations $\bm{H}'$ $\in\mathbb{R}^{N\times (d_{h}+d_{f})} =[\bm{H}, \bm{\tilde{H}}\bm{W}_2]$, where $\bm{W}_2\in\mathbb{R}^{d_{h}\times d_{f}}$ indicates a linear transformation for dimensionality reduction.\footnotemark[4]

\footnotetext[4]{We employ third-party tools to perform parsing for the corpora that do not contain gold syntax annotations. Since sometimes parsing may fail, dependency-guided GCN will be noneffective. Concatenation can remedy such problem since $\bm{H}$ still works even if $\bm{\tilde{H}}$ is invalid.}

\subsection{Span Representation}
\label{sec:span_rep}
We employ span enumeration~\cite{luan2019general} to generate text spans. Take the sentence ``The mitral valve leaflets are mildly thickened'' in Figure~\ref{fig:model} as an example, the generated text spans will be ``The'', ``The mitral'', ``The mitral valve'', ..., ``mildly'', ``mildly thickened'' and ``thickened''. To represent a text span, we use the concatenation of word representations of its startpoint and endpoint. For example, given word representations $\bm{H}$ = \{$\bm{h}_1$, $\bm{h}_2$, ..., $\bm{h}_N$\} $\in\mathbb{R}^{N\times d_{h}}$ (or $\bm{H}'$ = \{$\bm{h}'_1$, $\bm{h}'_2$, ..., $\bm{h}'_N$\}) and a span $(i,j)$ that starts at the position $i$ and ends at $j$, the span representation will be
\begin{equation}
\bm{s}_{i,j} = [\bm{h}_i, \bm{h}_j, \bm{w}]\ \mathrm{or}\  [\bm{h}'_i, \bm{h}'_j, \bm{w}],
\end{equation}
where $\bm{w}$ is a 20-dimensional embedding to represent the span width following previous work~\cite{luan2019general,wadden2019entity}. Thus, the dimension $d_s$ of $\bm{s}_{i,j}$ is $2d_{h}+20$ (or $2(d_{h}+d_{f})+20$).

\subsection{Decoding}
\label{sec:decode}

Our decoding consists of two parts.
First, we recognize all valid entity fragments,
and then perform pairwise classifications over the fragments to uncover their relationships.

\noindent\textbf{Entity Fragment Recognition:} Given a span $(i,j)$ represented as $\bm{s}_{i,j}$, we utilize one MLP to classify whether the span is an entity fragment and what is the entity type, formalized as:
\begin{equation} \label{eq:ner}
  \bm{p}_1 = \mathrm{softmax}(\mathrm{MLP}_1(\bm{s}_{i,j})),
\end{equation}
where $\bm{p}_1$ indicates the probabilities of entity types such as \emph{Organization}, \emph{Disease} and \emph{None} (i.e., not an entity fragment).

\noindent\textbf{Fragment Relation Prediction:}
Given two entity fragments $(i,j)$ and $(\tilde{i},\tilde{j})$ represented as $\bm{s}_{i,j}$ and $\bm{s}_{\tilde{i},\tilde{j}}$, we utilize another MLP to classify their relations:
\begin{equation} \label{eq:re}
  \bm{p}_2 = \mathrm{softmax}(\mathrm{MLP}_2([\bm{s}_{i,j}, \bm{s}_{i,j}*\bm{s}_{\tilde{i},\tilde{j}}, \bm{s}_{\tilde{i},\tilde{j}}])),
\end{equation}
where $\bm{p}_2$ indicates the probabilities of three classes, namely \emph{Succession}, \emph{Overlapping} and \emph{Other},
and the feature representations are mostly referred from \citet{luan2019general} and \citet{wadden2019entity}.
Noticeably, although the overlapped entities can be recognized at the first step,
here we use the \texttt{Overlapping} as one auxiliary strategy to further enhance the model.

\begin{algorithm}[t]
\small
\caption{\label{alg:decode} Decoding algorithm.}
\begin{algorithmic}[1]
    \REQUIRE An input sentence $\bm{x}$ = \{$x_1$, $x_2$, ..., $x_N$\}
    \ENSURE The recognized results $R$
        \STATE  $S$ = \textsc{EnumerateSpan}($\bm{x}$) where $S=\{s_{1,1}, s_{1,2}, ...\}$
        \FOR{$s_{i,j}$ in $S$}
            \IF{\textsc{IsEntityFragment}($s_{i,j}$)}
                \STATE  $V$ $\leftarrow$ $s_{i,j}$
            \ENDIF
        \ENDFOR
        \FOR{each pair $s_{i,j}$, $s_{\tilde{i},\tilde{j}}$ in $V$}
            \IF{\textsc{IsSuccession}($s_{i,j}$, $s_{\tilde{i},\tilde{j}}$)}
                \STATE  $E$ $\leftarrow$ $<s_{i,j}, s_{\tilde{i},\tilde{j}}>$
            \ENDIF
        \ENDFOR
        \STATE Graph $G=\{V, E\}$
        \FOR{$g$ in \textsc{FindCompleteSubgraphs}($G$)}
            \STATE  $R$ $\leftarrow$ $g$
        \ENDFOR
        \RETURN $R$
\end{algorithmic}
\end{algorithm}

\footnotetext[5]{We only use the \texttt{Succession} relations during decoding while ignore the \texttt{Overlapping} relations. The \texttt{Overlapping} relations are only used during training.}

\begin{table*}[t]
\small
\renewcommand{\arraystretch}{1.30}

\begin{center}
\begin{tabular}{|c|c|c|c|c|c|c|}
\hline
& & \bf CLEF & \bf CLEF-Dis & \bf CADEC & \bf GENIA & \bf ACE \\
\hline
\multirow{3}{*}{\shortstack{\# Documents or Sentences}} & Train & 179  & 534 & 875 & 1,599 & 370 \\
\cline{2-7}
& Dev & 20 & 303 & 187 & 200 & 43 \\
\cline{2-7}
& Test & 99 & 430 & 188 & 200 & 51 \\
\hline
\hline
\multirow{3}{*}{\shortstack{\% of Overlapped Entities}} & Train & 6  & 29 & 15 & 18 &  40 \\
\cline{2-7}
& Dev & 7 & 38 & 14 & 18 &  37 \\
\cline{2-7}
& Test & 8 & 36 & 13 & 22 &  39 \\
\hline
\hline
\multirow{3}{*}{\shortstack{\% of Discontinuous Entities}} & Train &  11 & 54 & 11 & 0 & 0 \\
\cline{2-7}
& Dev & 13 & 55 & 10 & 0 & 0 \\
\cline{2-7}
& Test & 8 & 52 & 9 & 0 & 0 \\
\hline
\end{tabular}
\end{center}
\caption{\label{tbl:datastat} Dataset statistics. For the CLEF, CLEF-Dis, CADEC, GENIA and ACE05 datasets, we follow the settings of \citet{dai-etal-2020-effective}, \citet{wang2019combining}, \citet{luan2019general} and \citet{lu2015joint} respectively. The statistics of CLEF-Dis are sentence numbers, others are document numbers. 
}
\end{table*}

During decoding (Algorithm \ref{alg:decode}), our model recognizes entity fragments from text spans (lines 2-4) in the input sentence and selects each pair of these fragments to determine their relations (lines 5-7).
Therefore, the prediction results can be considered as an \emph{entity fragment relation graph} (line 8), where a node denotes an entity fragment and an edge denotes the relation between two entity fragments.\footnotemark[5]
The decoding object is to find all the subgraphs in which each node connects with any other node (line 9). Thus, each of such subgraph composes an entity (line 10). 
In particular, the entity fragment that has no edge with others composes an entity by itself.

\subsection{Training}
\label{sec:train}

During training, we employ multi-task learning~\cite{caruana1997multitask,liu2017adversarial} to jointly train different parts of our model.\footnotemark[6] The loss function is defined as the negative log-likelihood of the two classification tasks, namely \emph{Entity Fragment Recognition} and \emph{Fragment Relation Prediction}:
\begin{equation} \label{eq:loss}
  \mathcal{L} = -\sum \alpha \log\bm{p}_1(y_{\text{ent}}) + \beta \log\bm{p}_2(y_{\text{rel}}),
\end{equation}
where $y_{\text{ent}}$ and $y_{\text{rel}}$ denote the corresponding gold-standard labels for text spans and span pairs,
$\alpha$ and $\beta$ are the weights to control the task importance.
During training, we use the BertAdam algorithm~\cite{devlin-etal-2019-bert} with the learning rate $5\times10^{-5}$ to finetune BERT and $1\times10^{-3}$ to finetune other parts of our model. The training process would terminate if the performance does not increase by 15 epochs. 

\footnotetext[6]{Please refer to Appendix \ref{app:effect_joint} for the effect of multi-task learning.}

\section{Experimental Setup}

\noindent \textbf{Datasets:} To evaluate our model for simultaneously recognizing overlapped and discontinuous entities, we follow prior work~\cite{muis2016learning,wang2019combining,dai-etal-2020-effective} and employ the data, called \textbf{CLEF}, from the ShARe/CLEF eHealth Evaluation Lab 2013~\cite{suominen2013overview}, which consists of 199 and 99 clinical notes for training and testing.
Note that \citet{dai-etal-2020-effective} used the full CLEF dataset in their experiments (179 for training, 20 for development and 99 for testing), while \citet{muis2016learning} and \citet{wang2019combining} used a subset of the union of the CLEF dataset and SemEval 2014 Task 7~\cite{pradhan2014semeval}.
Concretely, they used the training set and test set of the ShARe/CLEF eHealth Evaluation Lab 2013 as the training and development set, and they also used the development set of the SemEval 2014 Task 7 as the test set. In addition, they selected only the sentences that contain at least one discontinuous entity. Finally, the training, development and test sets contain 534, 303 and 430 sentences, respectively. We call this dataset as \textbf{CLEF-Dis} in this paper.
Moreover, we also follow \citet{dai-etal-2020-effective} to evaluate models using the \textbf{CADEC} dataset proposed by \citet{karimi2015cadec}. We follow the setting of \citet{dai-etal-2020-effective} to split the dataset and conduct experiments.

To show our model is comparable with the state-of-the-art models for overlapped NER, we conduct experiments on \textbf{GENIA}~\cite{kim2003genia} and \textbf{ACE05}. For the GENIA and ACE05 datasets, we employ the same experimental setting in previous works~\cite{lu2015joint,muis2017labeling,wang2018neural,luan2019general}, where 80\%, 10\% and 10\% sentences in 1,999 GENIA documents, and the sentences in 370, 43 and 51 ACE05 documents are used for training, development and test, respectively. The statistics of all the datasets we use in this paper is shown in Table~\ref{tbl:datastat}.

\noindent \textbf{Evaluation Metrics:} In terms of evaluation metrics, we follow prior work~\cite{lu2015joint,muis2016learning,wang2018neural,wang2019combining} and employ the precision (P), recall (R) and F1-score (F1). A predicted entity is counted as true-positive if its boundary and type match those of a gold entity. For a discontinuous entity, each span should match a span of the gold entity.
All F1 scores reported in Section~\ref{sec:result_analyze} are the mean values from five runs of the same setting.

\noindent \textbf{Implementation Details:} For hyper-parameters and other details, please refer to Appendix \ref{app:imp_detail}.

\section{Results and Analyses}
\label{sec:result_analyze}

\subsection{Results on CLEF}
\label{sec:result_clef}

\begin{table}[t]
\small
\renewcommand{\arraystretch}{1.30}

\begin{center}
\begin{tabular}{|c|l|c|}
\hline
\bf Related Work & \bf Method  & \bf F1 \\
\hline
\citet{tang2013recognizing} & CRF, BIOHD &  75.0 \\
\citet{tang2015recognizing} & CRF, BIOHD1234 &  78.3 \\
\citet{dai-etal-2020-effective} & Transition-based, ELMo\footnotemark[7] & 77.7\\
\hline
\multirow{4}{*}{Our Model} & Span-based, BERT & \textbf{83.2}\\
    & {\color{white}0}   -- Dep-guided GCN &  82.5\\
 & {\color{white}0}   -- Overlap Relation & 82.2\\
  & {\color{white}0}   -- BERT & 78.6 \\
\hline
\end{tabular}
\end{center}
\caption{\label{tbl:dis-ner-whole} Results on the CLEF dataset. }
\end{table}

\begin{table}[t]
\small
\renewcommand{\arraystretch}{1.30}

\begin{center}
\begin{tabular}{|c|l|c|}
\hline
\bf Related Work & \bf Method & \bf F1 \\
\hline
\citet{muis2016learning} & Hypergraph & 52.8\\
\citet{wang2019combining} & Hypergraph, RNN & 56.1\\
\citet{dai-etal-2020-effective} & Transition-based, ELMo &  62.9\\
\hline
\multirow{4}{*}{Our Model} & Span-based, BERT & \textbf{63.3}\\
    & {\color{white}0}   -- Dep-guided GCN &  62.9\\
 & {\color{white}0}   -- Overlap Relation & 62.6\\
  & {\color{white}0}   -- BERT & 56.4 \\
\hline
\end{tabular}
\end{center}
\caption{\label{tbl:dis-ner} Results on the CLEF-Dis dataset.}
\end{table}

Table~\ref{tbl:dis-ner-whole} shows the results on the CLEF dataset. As seen, \citet{tang2013recognizing} and \citet{tang2015recognizing} adapted the CRF model, which is usually used for flat NER, to overlapped and discontinuous NER. They modified the \texttt{BIO} label scheme to \texttt{BIOHD} and \texttt{BIOHD1234}, which use ``H'' to label overlapped entity segments and ``D'' to label discontinuous entity segments. Surprisingly, the recently-proposed transition-based model \cite{dai-etal-2020-effective} does not perform better than the CRF model \cite{tang2015recognizing}, which may be because \citet{tang2015recognizing} have conducted elaborate feature engineering for their model.
In contrast, our model outperforms all the strong baselines with at least about 5\% margin in F1. Our model does not rely on feature engineering or manually-designed transitions, which is more suitable for modern end-to-end learning.
\footnotetext[7]{\citet{dai-etal-2020-effective} found that BERT did not perform better than ELMo in their experiments.}

We further perform ablation studies to investigate the effect of dependency-guided GCN and the overlapping relation,
which can be removed without influencing our major goal.
As shown in Table~\ref{tbl:dis-ner-whole}, after removing either of them, the F1 scores go down by 0.7\% and 1.0\%.
The observation suggests that both dependency-guided GCN and the overlapping relation are effective for our model.
Moreover, after we replace BERT with the word embeddings pretrained on PubMed \cite{chiu2016train}, the F1 score goes down by 4.6\%, which demonstrates that BERT plays an important role in our model.

\subsection{Results on CLEF-Dis}
\label{sec:result_clef_dis}

Table~\ref{tbl:dis-ner} shows the results on the CLEF-Dis dataset.
As seen, our model outperforms the previous best model \cite{dai-etal-2020-effective} by 0.4\% in F1, which indicates that our model is very competitive,
leading to a new state-of-the-art result on the dataset.
Similarly, we further perform ablation studies to investigate the effect of dependency-guided GCN, the overlapping relation and BERT on this dataset.
As shown, after removing either of the GCN or overlapping relation, the F1 score decreases by 0.4\% or 0.7\%, which is consistent with the observations in Table~\ref{tbl:dis-ner-whole}.
In addition, to fairly compare with \citet{wang2019combining}, we also replace BERT with the word embeddings pretrained on PubMed \cite{chiu2016train}.
As we can see, our model also outperforms their model by 0.3\%.

\subsection{Results on CADEC}
\label{sec:result_cadec}

\begin{table}[t]
\small
\renewcommand{\arraystretch}{1.30}

\begin{center}
\begin{tabular}{|c|l|c|}
\hline
\bf Related Work & \bf Method & \bf F1 \\
\hline
Baseline (\citeyear{metke2016concept}) & CRF, BIOHD & 60.2\\
\citet{tang2018recognizing} & LSTM-CRF, Multilabel & 66.3\\
\citet{dai-etal-2020-effective} & Transition-based, ELMo &  69.0\\
\hline
\multirow{4}{*}{Our Model} & Span-based, BERT & 69.5\\
    & {\color{white}0}   -- Dep-guided GCN &  \textbf{69.9}\\
 & {\color{white}0}   -- Overlap Relation & \textbf{69.9}\\
  & {\color{white}0}   -- BERT & 66.8 \\
\hline
\end{tabular}
\end{center}
\caption{\label{tbl:result_cadec} Results on the CADEC dataset. ``Baseline (\citeyear{metke2016concept})'' indicates \citet{metke2016concept}.}
\end{table}

As shown in Table \ref{tbl:result_cadec}, \citet{metke2016concept} employed the similar method in \cite{tang2013recognizing} by expanding the \texttt{BIO} label scheme to \texttt{BIOHD}. \citet{tang2018recognizing} also experimented the \texttt{BIOHD} label scheme, but they found that the result of the \texttt{BIOHD}-based method was slightly worse than that of the ``Multilabel'' method (65.5\% vs. 66.3\% in F1). 
Compared with the method in \cite{metke2016concept}, the performance improvement might be mainly because they used deep neural networks (e.g., LSTM) instead of shallow non-neural models.

Compared with the above baselines, the transition-based model \citet{dai-etal-2020-effective} is still the best. Our full model slightly outperforms the transition-based model by 0.5\%. In this dataset, we do not observe mutual benefit between the dependency-guided GCN and overlapped relation prediction modules, since our model achieves better results when using them separately (69.9\%) than using them jointly (69.5\%).
However, when using them separately, the F1 is still 0.6\% higher than the one using neither of them.
Without BERT, the performance of our model drops by about 3\% but it is still comparable with the performances of the methods without contextualized representations.

\subsection{Result Analysis based on Entity Types}

\begin{figure}[t]
\centering
\includegraphics[scale=0.95]{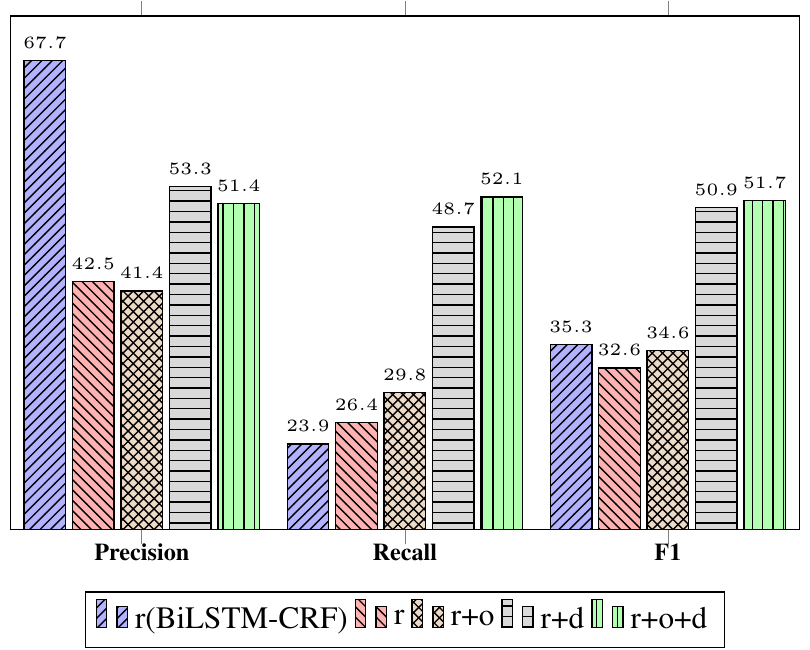}
\caption{\label{fig:rod} Result analysis based on entity types (i.e., (r)egular, (o)verlapped and (d)iscontinuous) on the CLEF-Dis dataset, comparing with BiLSTM-CRF.
\footnotemark[8]
}
\end{figure}

\paragraph{Comparing with BiLSTM-CRF}
To show the necessity of building one model to recognize regular, overlapped and discontinuous entities simultaneously, we analyze the predicted entities in the CLEF-Dis dataset and classify them based on their types, as shown in Figure~\ref{fig:rod}. 
In addition, we compare our model with BiLSTM-CRF \cite{lample-EtAl:2016:N16-1,ma-hovy:2016:P16-1,yang2018design}, to show our model does not influence the performance of regular NER significantly. 
For a fair comparison, we replace BERT with Glove \cite{pennington2014glove} and keep the setting of our model the same with the setting of the BiLSTM-CRF model used in previous work~\cite{yang2018design}.

\footnotetext[8]{Many discontinuous entities are also overlapped, but we do not count them as overlapped entities in this figure.}

\begin{figure}[t]
\centering
\includegraphics[scale=0.95]{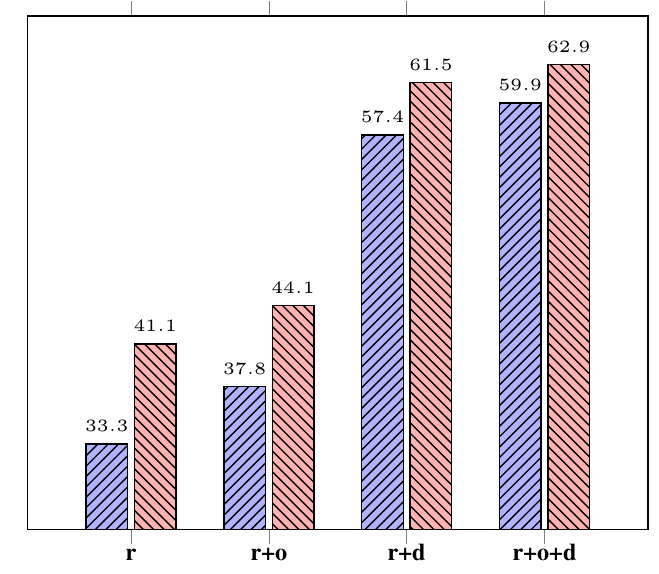}
\caption{\label{fig:rod_dai} Result analysis based on entity types on the CLEF-Dis dataset, comparing with \citet{dai-etal-2020-effective} (blue).
}
\end{figure}

\begin{table*}[t]
\small
\renewcommand{\arraystretch}{1.30}

\begin{center}
\begin{tabular}{|c|l|c|c|}
\hline
 \bf Related Work & \bf Method & \bf GENIA & \bf ACE05 \\
\hline
\citet{finkel2009nested} & Constituency parsing & 70.3 & \textendash \\
\citet{lu2015joint} & Hypergraph & 70.3 & 58.7 \\
\citet{muis2017labeling} & Hypergraph & 70.8 & 61.3 \\
\citet{katiyar2018nested} & Hypergraph, RNN & 73.8 & 70.5 \\
\citet{wang2018transition} & Transition-based parsing, RNN  & 73.9 & 73.0 \\
\citet{ju2018neural} & Dynamically stacking, RNN & 74.7 & 72.2 \\
\citet{zheng2019boundary} & Boundary detection, RNN & 74.7 & \textendash \\
\citet{lin2019sequence} & Anchor-region detection, RNN, CNN & 74.8 & 74.9 \\
\citet{wang2018neural} & Hypergraph, RNN  & 75.1 & 74.5 \\
\citet{xia2019multi} & Multi-grained detection, RNN, ELMo & \textendash & 78.2 \\
\citet{fisher2019merge} & Merge and label, BERT & \textendash & 82.4 \\
\citet{luan2019general} & Span-based, ELMo, Coref   & 76.2 & 82.9 \\
\citet{wadden2019entity} & Span-based, BERT, Coref  & 77.9 & \textendash \\
\citet{strakova2019neural} & Seq2Seq, ELMo, BERT, Flair  & \textbf{78.3} & \textbf{84.3} \\
\hline
\multirow{3}{*}{Our Model} & Span-based, BERT & 77.8 & 83.0\\
    & {\color{white}0}   -- Dep-guided GCN &  77.4 & 82.6\\
 & {\color{white}0}   -- Overlap Relation & 77.4 & 82.7\\
\hline
\end{tabular}
\end{center}
\caption{\label{tbl:nested-ner} Comparisons with prior work on the GENIA and ACE05 datasets. }
\end{table*}

As seen, if only considering regular entities, the BiLSTM-CRF model can achieve a better performance compared with our model,
especially the precision value is much higher.
One likely reason might be that the BiLSTM-CRF model is capable of using the label dependence to detect entity boundaries accurately,
ensuring the correctness of the recognized entities, which is closely related to the precision.
Nevertheless, our model can lead to higher recall,
which reduces the gap between the two models.

If considering both regular and overlapped entities, the recall of our model is greatly boosted, and thus the F1 increases concurrently.
If both regular and discontinuous entities are included, the performance of our model rises significantly to 50.9\% due to the large scale of discontinuous entities.
When all types of entities are concerned, the F1 of our model further increases by 0.8\%,
indicating the effectiveness of our model in joint recognition of overlapped, discontinuous and regular entities.

\paragraph{Comparing with the Transition-Based Model}
As shown in Figure \ref{fig:rod_dai}, we also compare our model with the transition-based model \cite{dai-etal-2020-effective} based on entity types by analyzing the results from one run of experiments. Note that since we do not tune the hyper-parameters of the transition-based model elaborately, the performance is not as good as the one that they have reported. As seen, our model performs better in all of the four groups, namely regular, regular+overlapped, regular+discontinuous, regular+overlapped+discontinuous entity recognition. 
However, based on the observation on the bars in different groups, we find that the main superiority of our model comes from regular entity recognition.
In recognizing overlapped entities, our model is comparable with the transition-based model, but in recognizing discontinuous entities, our model performs slightly worse than the transition-based model. 
This suggests that a combination of span-based and transition-based models may be a potential method for future research.

\subsection{Results on GENIA and ACE05}
\label{sec:result_genia}

Table~\ref{tbl:nested-ner} shows the results of the GENIA and ACE05 datasets, which include only regular and overlapped entities.
Our final model achieves 77.8\% and 83.0\% F1s in the GENIA and ACE05 datasets, respectively.
By removing the dependency-guided GCN, the model shows an averaged decrease of 0.4\%,
indicating the usefulness of dependency syntax information.
The finding is consistent with that of the CLEF dataset.
Interestingly, we note that the overlapping relation also brings a positive influence in this setting.
Actually, the relation extraction architecture is not necessary for only regular and overlapped entities,
because the decoding can be finished after the first entity fragment recognition step.
The observation doubly demonstrates the advantage of our final model.
We also compare our results with several state-of-the-art results of the previous work on the two datasets in Table \ref{tbl:nested-ner}.
Only the studies with the same training, development and test divisions are listed.
We can see that our model can achieve very competitive performances on both datasets.
Note that \citet{luan2019general} and \citet{wadden2019entity} use extra coreference resolution information,
and \citet{strakova2019neural} exploit much richer word representations by a combination of ELMo, BERT and Flair.

\section{Conclusion}
In this work, we proposed an efficient and effective model to recognize both overlapped and discontinuous entities simultaneously,
which can be applied to any NER dataset theoretically, since no extra assumption is required to limit the type of named entities.
First, we enumerate all spans in a given sentence to determine whether they are valid entity fragments,
and then relation classifications are performed to check the relationships between all fragment pairs.
The results show that our model is highly competitive to the state-of-the-art models for overlapped or discontinuous NER.
We have conducted detailed studies to help comprehensive understanding of our model.

\section*{Acknowledgments}

We thank the reviewers for their comments and recommendation.
This work is supported by the National Natural Science Foundation of China (No. 61772378), 
the National Key Research and Development Program of China (No. 2017YFC1200500), 
the Research Foundation of Ministry of Education of China (No. 18JZD015).

\bibliographystyle{acl_natbib}
\bibliography{acl2021}

\clearpage

\appendix

\section{Comparing Different Settings in the Word Representation Layer}
\label{app:cmp_wordrep}

\begin{table}[t]
\small
\renewcommand{\arraystretch}{1.30}

\begin{center}
\begin{tabular}{|l|c|c|}
\hline
\bf Method & \bf CLEF & \bf CLEF-Dis \\
\hline
Word2Vec & 68.5 & 43.5 \\
Word2Vec+BiLSTM & 78.6 &  56.4 \\
ELMo & 74.2 & 48.1 \\
ELMo+BiLSTM & 77.1 &  55.8 \\
BERT & 82.5 &  59.0 \\ 
BERT+BiLSTM & 83.2 &  63.3 \\
\hline
\end{tabular}
\end{center}
\caption{\label{tbl:cmp_wordrep} Results using different word representation methods.}
\end{table}

The word representation layer addresses the problem that how to transform a word into a vector for the usage of upper layers. In this paper, we investigate several common word encoders in recent NLP research to generate word representations, namely Word2Vec~\cite{mikolov2013word2vec} (or its variants such as Glove~\cite{pennington2014glove}), ELMo~\cite{peters2018deep} and BERT~\cite{devlin-etal-2019-bert}. 
Given an input sentence $\bm{x}$ = \{$x_1$, $x_2$, ..., $x_N$\}, we use different methods to represent them as vectors based on which word encoders we utilize:
\begin{itemize}
  \item If Word2Vec is used, each word $x_i$ will be directly transformed into a vector $\bm{h}_i$ according to the pretrained embedding lookup table. Therefore, all the words in the sentence $\bm{x}$ correspond to a matrix $\bm{H}$ = \{$\bm{h}_1$, $\bm{h}_2$, ..., $\bm{h}_N$\} $\in\mathbb{R}^{N\times d_{h}}$, where $d_{h}$ denotes the dimension of $\bm{h}_i$.
  \item If ELMo is used, each word $x_i$ will first be split into characters and then input into character-level convolutional networks to obtain character-level word representations. Finally, all word representations in the sentence will be input into 3-layer BiLSTMs to generate contextualized word representations, which can also be denoted as $\bm{H}$ = \{$\bm{h}_1$, $\bm{h}_2$, ..., $\bm{h}_N$\}
  \item If BERT is used, each word $x_i$ will be converted into word pieces and then fed into a pretrained BERT module. After the BERT calculation, each sentential word may involve vectorial representations of several pieces. Here we employ the representation of the beginning word piece as the final word representation following \cite{wadden2019entity}. For instance, if ``fevers'' is split into ``fever'' and ``\#\#s'', the representation of ``fever'' is used as the whole word representation. Therefore, all the words in the sentence $\bm{x}$ can also be represented as a matrix $\bm{H}$ = \{$\bm{h}_1$, $\bm{h}_2$, ..., $\bm{h}_N$\}
\end{itemize}
In addition, a bidirectional LSTM (BiLSTM) layer can be stacked on word encoders to further capture contextual information in the sentence, which is especially helpful for non-contextualized word representations such as Word2Vec. Concretely, the word representations $\bm{H}$ = \{$\bm{h}_1$, $\bm{h}_2$, ..., $\bm{h}_N$\} will be input into the BiLSTM layer and consumed in the forward and backward orders. Assuming that the outputs of the forward and backward LSTMs are $\overrightarrow{\bm{H}}$ = \{$\overrightarrow{\bm{h}}_1$, $\overrightarrow{\bm{h}}_2$, ..., $\overrightarrow{\bm{h}}_N$\} and $\overleftarrow{\bm{H}}$ = \{$\overleftarrow{\bm{h}}_1$, $\overleftarrow{\bm{h}}_2$, ..., $\overleftarrow{\bm{h}}_N$\} respectively. Thus, they can be concatenated (e.g., $\hat{\bm{h}}_i=[\overrightarrow{\bm{h}}_i, \overleftarrow{\bm{h}}_i]$) to compose the final word representations $\hat{\bm{H}}$ = \{$\hat{\bm{h}}_1$, $\hat{\bm{h}}_2$, ..., $\hat{\bm{h}}_N$\}.

\begin{table*}[t]
  \centering
  \small
  \renewcommand{\arraystretch}{1.3}
  
  \begin{tabular}{|p{6cm}|C{9cm}|}
   \hline
    \multicolumn{1}{|c|}{\textbf{Examples}} & \textbf{Dependency Graphs} \\
    \hline
    This showed a mildly $[\textbf{displaced}]_1$ and  & \multirow{3}{*}{\includegraphics[scale=0.5]{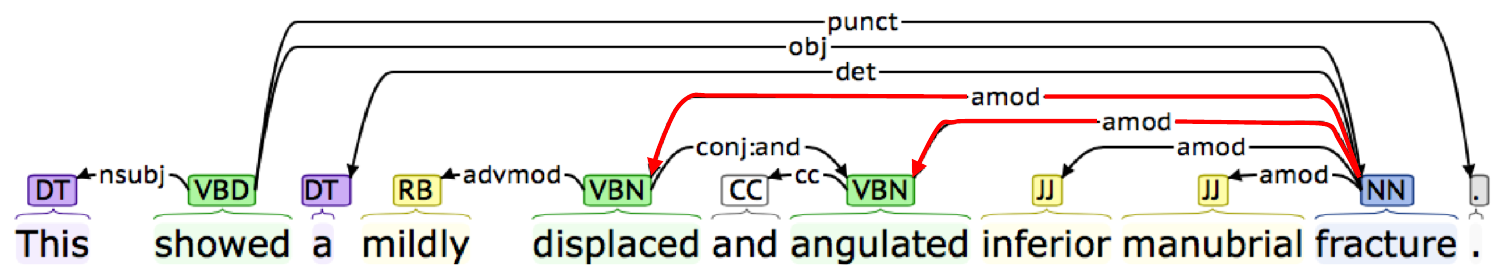}} \\
    $[\textbf{angulated}]_2$ inferior manubrial $[[\textbf{fracture}]_1]_2$. & \\
    & \\
    \hline
    $[[\textbf{Tone}]_1]_2$ was $[\textbf{increased}]_1$ in the left lower  & \multirow{4}{*}{\includegraphics[scale=0.5]{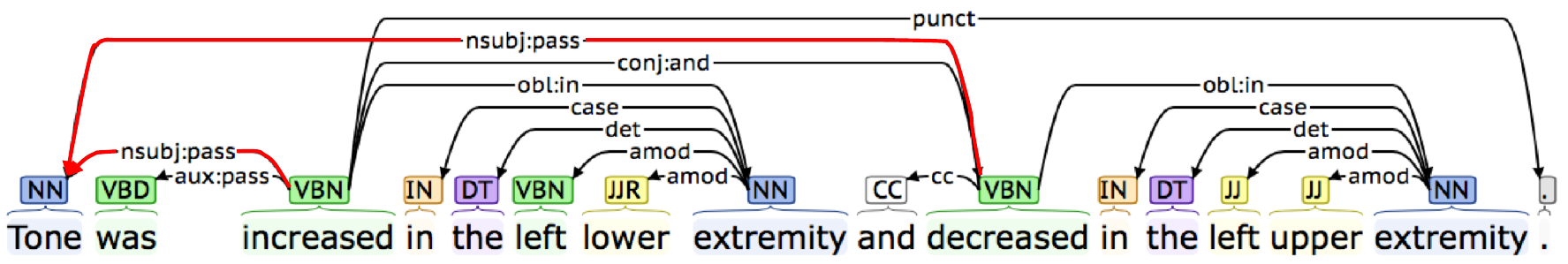}} \\
    extremity and $[\textbf{decreased}]_2$ in the left upper  & \\
    extremity. & \\
    & \\
    \hline
  \end{tabular}
  \caption{\label{tbl:case} Case Studies. Bold words with the same number belong to the same entity.}
\end{table*}

We investigate the effects of different word encoders and the BiLSTM layer in the experiments. As shown in Table \ref{tbl:cmp_wordrep}, we compare the effects of different word representation methods in the CLEF and CLEF-Dis datasets, where the size of the former one is much bigger than that of the latter, in order to also investigate the impact of the data size on word representations. From the table, the first observation is that BERT is the most effective word representation method. Surprisingly, Word2Vec is more effective than ELMo, which may be because ELMo is exclusively based on characters and cannot effectively capture the whole meanings of words. Therefore, this suggests that it is better to use ELMo with Word2Vec.

Second, we find that BiLSTM is helpful in all cases, especially for Word2Vec. This may be because Word2Vec is a kind of non-contexualized word representations, which particularly needs the help of BiLSTM to capture contexual information. In contrast, BERT is not very sensitive to the help of BiLSTM as Word2Vec and ELMo, which may be because the transformer in BERT has already captured contexual information.

Third, we observe that the effect of BiLSTM is more obvious for the CLEF-Dis dataset. Considering the data sizes of the CLEF and CLEF-Dis datasets, it is more likely that small datasets need the help of BiLSTM, while big datasets are less sensitive to the BiLSTM and BERT is usually enough for them to build word representations.

\section{Case Studies}
\label{app:case_studies}

To understand how syntax information helps our model to identify discontinuous or overlapped entities,
we offer two examples in the CLEF dataset for illustration, as shown in Table~\ref{tbl:case}.
Both the two examples are failed in the model without using dependency information,
but are correctly recognized in our final model.
In the first example, the fragments ``displaced'' and ``fracture'' of the same entity are far away from each other in the original sentence,
while they are directly connected in the dependency graph.
Similarly, in the second example, the distance between ``Tone'' and ``decreased'' is 9 in the sentence, while their dependency distance is only 1.
These dependency connections can be directly modeled in dependency-guided GCN,
thus, resulting in strong clues for the NER,
which makes our final model work.

\section{Effect of Joint Training}
\label{app:effect_joint}

\begin{table}[t]
\small
\renewcommand{\arraystretch}{1.30}

\begin{center}
\begin{tabular}{|l|c|c|c|}
\hline
\bf Method & \bf P & \bf R & \bf F1 \\
\hline
EFR & 81.2 & 79.6 & 80.4\\
EFR(+FRP) & 81.4 & 80.1 & 80.7\\
\hline
\end{tabular}
\end{center}
\caption{\label{tbl:mtl} Effect of joint training between entity fragment recognition (EFR) and fragment relation prediction (FRP) on the CLEF-Dis dataset. P, R and F1 are the results for EFR.
}

\end{table}

As mentioned in Section 3.5, we employ multi-task learning to jointly train our model between two tasks, namely entity fragment recognition and fragment relation prediction.
Therefore, it is interesting to show the effect of joint training by observing the performance changes of the entity fragment recognition (EFR) task before and after adding the fragment relation prediction (FRP) task.
As seen in Table~\ref{tbl:mtl}, the F1 of entity fragment recognition increases by 0.3\% after adding the FRP task, which shows that the FRP task could improve the EFR task.
This suggests that the interaction between entity fragment recognition and fragment relation prediction could benefit our model,
which also indicates that end-to-end modeling is more desirable.

\section{Implementation Details}
\label{app:imp_detail}

\begin{table}[t]
\small
\renewcommand{\arraystretch}{1.30}

\begin{center}
\begin{tabular}{|c|c|c|c|c|}
\hline
 & \bf CLEF & \bf CADEC & \bf GENIA & \bf ACE05 \\
\hline
$d_h$ & 400 & 400 & 768 & 768 \\
\hline
$N_{head}$ & 4 & 4 & 4 & 4 \\
\hline
$l$ & 2 & 2 & 2 & 1 \\
\hline
$d_f$ & 20 & 20 & 64 & 64 \\
\hline
$d_s$ & 860 & 860 & 1,684 & 1,684 \\
\hline
MLP Layer & 1 & 1 & 2 & 2 \\
\hline
MLP Size & 150 & 150 & 150 & 150\\
\hline
$\alpha$ & 1.0 & 1.0 & 1.0 & 1.0 \\
\hline
$\beta$ & 1.0 & 1.0 & 1.0 & 0.6 \\
\hline
\end{tabular}
\end{center}
\caption{\label{tbl:para} Main hyper-parameter settings in our model for all the datasets. $d_h$--Section 3.1; $N_{head}$, $l$ and $d_f$--Section 3.2; $d_s$--Section 3.3; $\alpha$ and $\beta$--Section 3.5. Note that the hyper-parameter settings in the CLEF-Dis dataset is the same as those in the CLEF dataset.}
\end{table}

Our model is implemented based on AllenNLP~\cite{Gardner2017AllenNLP}. The number of parameters is about 117M plus BERT. We use one GPU of NVIDIA Tesla V100 to train the model, which occupies about 10GB memories. The training time for one epoch is between 2$\sim$6 minutes on different datasets.

Table~\ref{tbl:para} shows the main hyper-parameter values in our model. We tune the hyper-parameters based on the results of about 5 trials on development sets.
Below are the ranges tried for the hyper-parameters:
the GCN layer $l$ (1, 2), the GCN head $N_{head}$ (2, 4), the GCN output size $d_f$ (20, 48, 64), the MLP layer (1, 2), the MLP size (100, 150, 200), the loss weight $\alpha$ and $\beta$ (0.6, 0.8, 1.0).
Since we employ the BERT$_{BASE}$, the dimension $d_h$ of word representations is 768 except in the CLEF and CADEC datasets, where we use a BiLSTM layer on top of BERT to obtain word representations since we observe performance improvements. We try 200 and 400 hidden units for the BiLSTM layer.

Considering the domains of the datasets, we employ clinical BERT\footnotemark[1]~\cite{alsentzer2019publicly}, SciBERT\footnotemark[2]~\cite{beltagy-etal-2019-scibert} and Google BERT\footnotemark[3]~\cite{devlin-etal-2019-bert} for the CLEF (and CADEC), GENIA and ACE05 datasets, respectively.
In addition, since our model needs syntax information for dependency-guided GCN, but the datasets do not contain gold syntax annotations, we utilize the Stanford CoreNLP toolkit~\cite{manning2014stanford} to perform dependency parsing.

\footnotetext[1]{https://github.com/EmilyAlsentzer/clinicalBERT}
\footnotetext[2]{https://github.com/allenai/scibert}
\footnotetext[3]{https://github.com/google-research/bert}

\end{document}